\documentclass[10pt,twocolumn,letterpaper]{article}

\usepackage{cvpr}              % To produce the CAMERA-READY version
\newcommand{\BigName}[1]{\textbf{#1}}
\definecolor{cvprblue}{rgb}{0.21,0.49,0.74}
\usepackage[pagebackref,breaklinks,colorlinks,allcolors=cvprblue]{hyperref}
\usepackage{multirow}
\usepackage{diagbox}
\usepackage{booktabs}
\usepackage{multirow}
\usepackage{xcolor}
\usepackage{algorithm}
\usepackage{algpseudocode}
\usepackage{listings}
\usepackage{xspace}
\usepackage{tcolorbox}
\tcbuselibrary{breakable}

\def\paperID{6187} % *** Enter the Paper ID here
\def\confName{CVPR}
\def\confYear{2026}
\usepackage{colortbl} %\cellcolor
\definecolor{lightblue}{HTML}{E8F0FE}
\newcommand{\PaperName}{Beyond Words and Pixels: A Benchmark for Implicit World Knowledge Reasoning in Generative Models\xspace}
\newcommand{\benchname}{PicWorld}
\newcommand{\agentname}{PW-Agent}
\newcommand{\scorename}{PW-Score}

\definecolor{PromptBack}{HTML}{FFF1B8}
\definecolor{PromptFrame}{HTML}{C9A100}
\definecolor{PromptTitle}{HTML}{000000}

\lstdefinestyle{promptlisting}{
  breaklines=true,
  breakatwhitespace=false,
  postbreak=\mbox{\textcolor{PromptFrame}{\(\hookrightarrow\)}\space},
  basicstyle=\ttfamily\footnotesize,
  columns=fullflexible,
  keepspaces=true,
  tabsize=2,
  showstringspaces=false
}

\usepackage{multicol}
\newtcolorbox{promptbox}[1][]{
  breakable,                 % 关键：支持跨页
  colback=PromptBack,
  colframe=PromptFrame,
  coltitle=PromptTitle,
  fonttitle=\bfseries,
  title=#1,
  boxrule=0.8pt,
  arc=2pt,
  left=6pt,right=6pt,top=6pt,bottom=6pt
}

\newtcolorbox{promptbox*}[1][]{
  breakable,
  enhanced jigsaw,
  colback=PromptBack,
  colframe=PromptFrame,
  coltitle=PromptTitle,
  fonttitle=\bfseries,
  title=#1,
  boxrule=0.8pt,
  arc=2pt,
  left=6pt,right=6pt,top=6pt,bottom=6pt,
}

\title{\raisebox{-0.35cm}{\includegraphics[width=1.5cm]{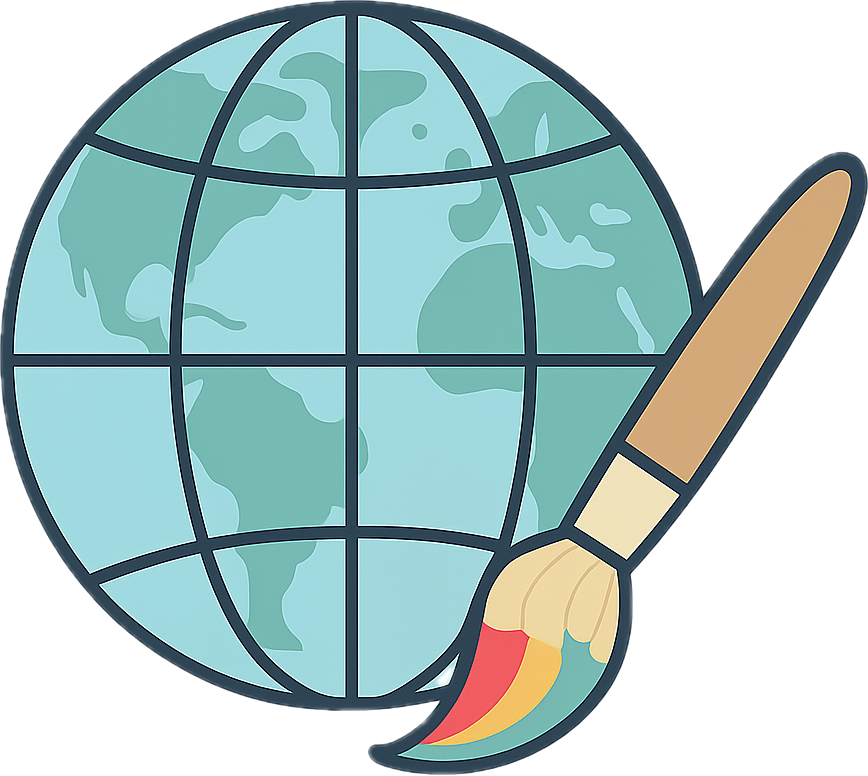}} \PaperName}
\author{\BigName{Tianyang Han}\textsuperscript{1,3}\thanks{Equal Contribution.} \quad
\BigName{Junhao Su}\textsuperscript{1}\footnotemark[1] \quad
\BigName{Junjie Hu}\textsuperscript{1,2}\footnotemark[1] \quad
\BigName{Kai Ma}\textsuperscript{1}\thanks{Project Leader.} \quad
\BigName{Peizhen Yang}\textsuperscript{4} \\
\BigName{Hengyu Shi}\textsuperscript{1} \quad
\BigName{Junfeng Luo}\textsuperscript{1} \quad
\BigName{Jialin Gao}\textsuperscript{1} \quad
\\
\textsuperscript{1}MeiGen AI Team, Meituan \quad
\textsuperscript{2}Fudan University \quad
\textsuperscript{3}$D^{4}$ Lab \\
\textsuperscript{4}HHMI Janelia Research Campus 
}

\begin{document}
\maketitle
\begin{abstract}
Text-to-image (T2I) models today are capable of producing photorealistic, instruction-following images, yet they still frequently fail on prompts that require implicit world knowledge. 
Existing evaluation protocols either emphasize compositional alignment or rely on single-round VQA-based scoring, leaving critical dimensions—such as knowledge grounding, multi-physics interactions, and auditable evidence-substantially under-tested.
To address these limitations, we introduce \benchname{}, the first comprehensive benchmark that assesses the grasp of implicit world knowledge and physical causal reasoning of T2I models. This benchmark consists of 1,100 prompts across three core categories. 
To facilitate fine-grained evaluation, we propose \agentname{}, an evidence-grounded multi-agent evaluator to hierarchically assess images on their physical realism and logical consistency by decomposing prompts into verifiable visual evidence. 
We conduct a thorough analysis of 17 mainstream T2I models on \benchname{},  illustrating that they universally exhibit a fundamental limitation in their capacity for implicit world knowledge and physical causal reasoning to varying degrees. 
The findings highlight the need for reasoning-aware, knowledge-integrative architectures in future T2I systems.

%By requiring per-fact, pixel-level evidence, our protocol mitigates the reliability and bias issues inherent in one-shot LLM-based scoring, while preserving the scalability benefits of query-driven evaluation.

\end{abstract}

\section{Introduction}
\label{sec:intro}

\begin{figure}[t]
    \centering
    \includegraphics[width=1.0\linewidth]{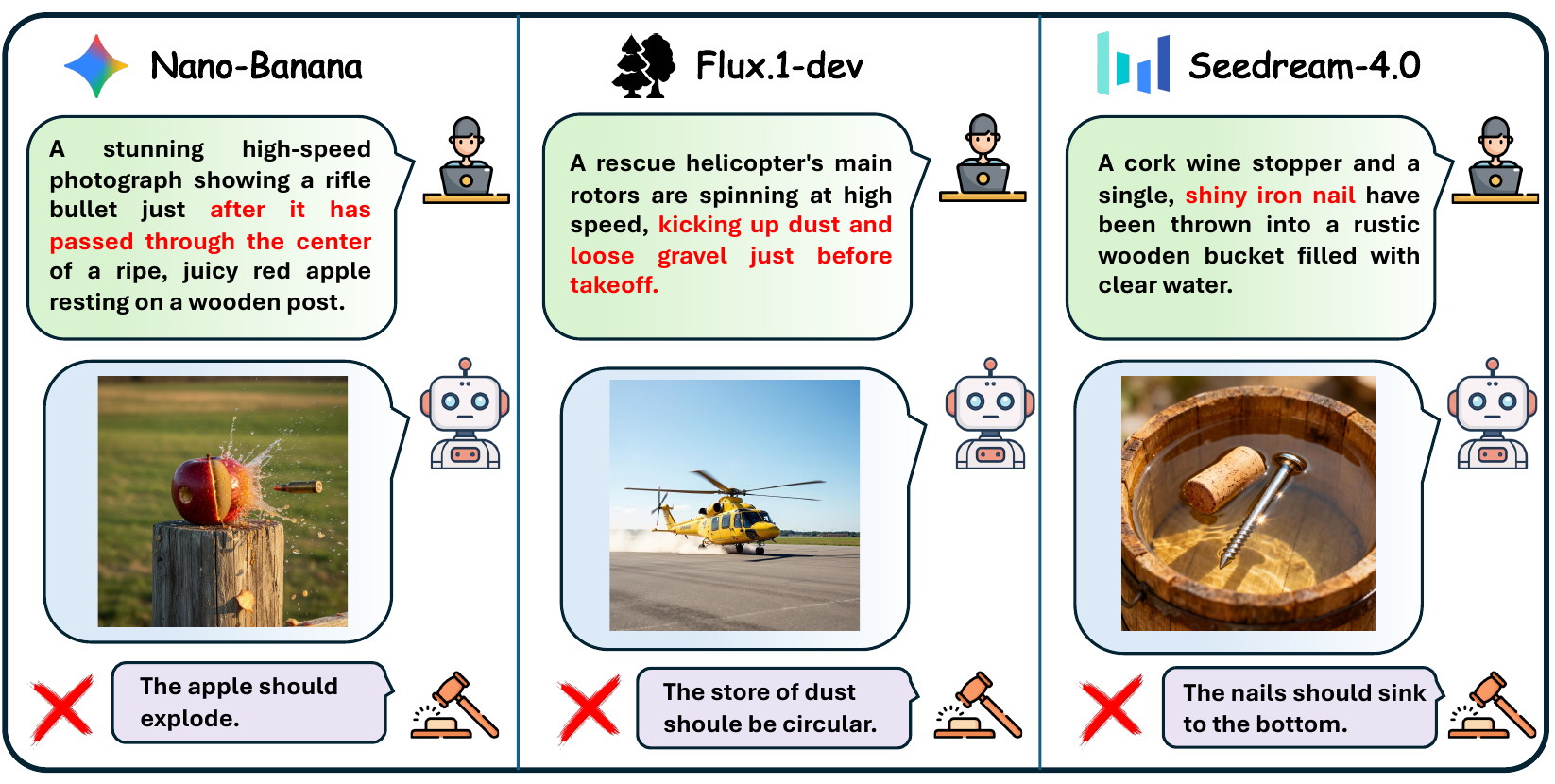}
    \caption{State-of-the-art T2I models often produce unrealistic images due to their lack of implicit world-knowledge reasoning.}
    \label{fig1}
    \vspace{-2mm}
\end{figure}

Text-to-image generation has rapidly advanced from producing plausible textures to synthesizing visually compelling, instruction-following scenes~\citep{flux, sd3, qwenimage, hu2025positionic, cai2025hidream, qin2025lumina}. However, an essential question remains under-explored: do current models truly ``understand'' the world they depict, the physical laws that govern appearances, the causal logic that binds actions to outcomes, and the fine-grained constraints embedded in complex prompts? As shown in Figure~\ref{fig1}, existing high-performing models still fail to reason and generate images consistent with world knowledge and natural laws, even though they can perfectly follow the given prompts~\citep{comanici2025gemini, seedream2025seedream}.

Conventional evaluation of text-to-image(T2I) models has centered on aesthetic quality and prompt fidelity using metrics~\citep{fid, hessel2021clipscore, rethinkingfid}. While recent benchmarks have started probing deeper reasoning, they often focus on specific aspects of cognition like explicit world-knowledge concepts and relational commonsense~\citep{phybench, commonsensebench, t2icompbench, zhang2025worldgenbench, niu2025wise}. 
These methods do not systematically test whether a model can generate the implied consequences of a scene rather than just its explicitly described components. 
For example, when prompted the model to generate ``a cork and an iron nail in a bucket of water'', the evaluation should not only confirm the objects'' presence but, more importantly, test its implicit understanding of buoyancy by checking if the nail sinks and the cork floats.
%Thus, a comprehensive evaluation that simultaneously probes implicit world knowledge, adherence to fundamental physical laws, and logical causal reasoning remains absent.
Thus, a comprehensive benchmark that simultaneously probes the implicit world knowledge, including adherence to fundamental physical laws, and logical causal reasoning of T2I models is urgently needed.

Furthermore, many recent T2I evaluation methods rely on a direct, zero-shot prompting approach by querying a powerful Multimodal Large Language Model (MLLM) as a holistic judger~\citep{llmasjudge, llmasjudges, schroeder2024reliability, su2025failure}. 
However, this single-round evaluation paradigm is not always reliable in producing accurate assessments, especially in the subjective task of scoring images. It may overlook consequential failures like an incorrect lateral inversion in a mirror, as its holistic assessment is not guided to assess for such specific physical laws. 
Similarly, it suffers from several critical issues inherent to current MLLMs, such as hallucinations~\citep{judgebias, han2024instinctive, bai2024hallucination}, where they may perceive objects or phenomena that are not actually present in the image, leading to inaccurate scores. 
Additionally, these models often exhibit a central tendency bias when faced with complex, multifaceted evaluation criteria. This behavior masks the true strengths and weaknesses of the model being evaluated.

To address the aforementioned issues, we introduce the \textbf{PicWorld} benchmark, the first comprehensive benchmark designed to systematically evaluate a text-to-image model's grasp of implicit world knowledge and physical reasoning. 
We meticulously leverage advanced MLLM Gemini-2.5-Pro to comprise 1,100 prompts organized into three core categories, Physical World, Abstract Knowledge and Logic\&Commensense Reasoning, followed by a rigorous manual curation process to ensure quality and clarity. 
Unlike previous method, \benchname{} provides a comprehensive framework for assessing a model's ability to adhere to fundamental physical laws and render logical causal relationships. 
It assesses the understanding of T2I models of implicit world knowledge and fundamental physical laws reasoning. It moves beyond semantic accuracy to test the logical consequences implied by a scene.

We also design an evidence-grounded, multi-agent pipeline \agentname{} to evaluate model performance on \benchname{}. The framework consists of four specialized agents: World Knowledge Extractor(WKE) that parses each prompt into atomic, image-verifiable expectations; Hypothesis Formulator(HF) that formulates verifiable visual questions; Visual Perceptor(VP) that grounds answers in visual evidence from the image; and Reasoning Judger(RJ) that aggregates answers via a deduction-based, continuous scoring scheme with checklist-style atomicity and importance weighting. RJ scores each image across three hierarchical dimensions: Instruction Adherence, Physical/Logical Realism, and Detail \& Nuance. This process yields a fine-grained, multi-faceted score for each generated image, providing a deep and explainable analysis of a model's reasoning capabilities.

We summarize our contributions as follows:
\begin{itemize}

    % \item We introduce \benchname{}, a comprehensive benchmark that evaluates T2I models’ implicit reasoning over world knowledge, physical laws, and causal logic.
    \item We introduce \benchname{}, a comprehensive benchmark designed to evaluate the implicit reasoning capabilities of text-to-image models. To the best of our knowledge, \benchname{} is the first large-scale, systematic benchmark specifically created to assess a model's understanding of implicit world knowledge like adherence to fundamental physical laws, and logical causal reasoning. 

    \item We propose \agentname{}, a novel, automated evaluation framework that employs Hierarchical Evaluation via Agentic Decomposition. This multi-agent pipeline systematically breaks down complex prompts into verifiable physical and logical components, enabling a reproducible and scalable analysis of model performance on our benchmark.

    \item  Our thorough experiments demonstrate that existing T2I models, particularly open-source ones, exhibit a limited capacity for physical and logical reasoning, highlighting key areas for future improvement.
\end{itemize}

\section{Related Work}
\label{sec:related}

\subsection{Advancements in Text-to-Image(T2I) Models}
Recent years, progress in T2I models enable the generation of high-quality images from natural language. This advancement has been propelled by several powerful architectural paradigms, such as diffusion models~\citep{flux, cai2025hidream, sd3, qin2025lumina, xie2024sana, chen2024pixart}, autoregressive models~\citep{wang2024emu3, chen2025janus, ma2025janusflow, var, chen2024high, han2025infinity} and multimodal unified models~\citep{bagel, xie2024show,chen2025janus, chen2025blip3,chen2025blip3o}. While these state-of-the-art models excel at creating visually impressive content, their outputs often reveal a superficial understanding of the world they depict.

\subsection{Evaluation of T2I Models}
Conventional evaluation of T2I models has primarily focused on aesthetic quality and prompt fidelity by metrics such as FID~\citep{fid} and CLIPScore~\citep{hessel2021clipscore}. 
Recent benchmarks have begun to probe the reasoning abilities of T2I systems. WorldGenBench~\citep{zhang2025worldgenbench} and WISE~\citep{niu2025wise} emphasize reasoning and generation grounded in world knowledge, while Commonsense-T2I~\citep{commonsensebench} and R2I-Bench~\citep{r2ibench} target broad commonsense knowledge and textual commonsense reasoning respectively.

Nevertheless, existing benchmarks either solely assess explicit prompt following or focus on only a partial view of model competence such as world-knowledge concepts, commonsense. A comprehensive evaluation that simultaneously probes implicit world knowledge reasoning such as adherence to fundamental physical laws, and logical causal reasoning remains absent.

\begin{figure*}[t]
    \centering
    \includegraphics[width=\textwidth]{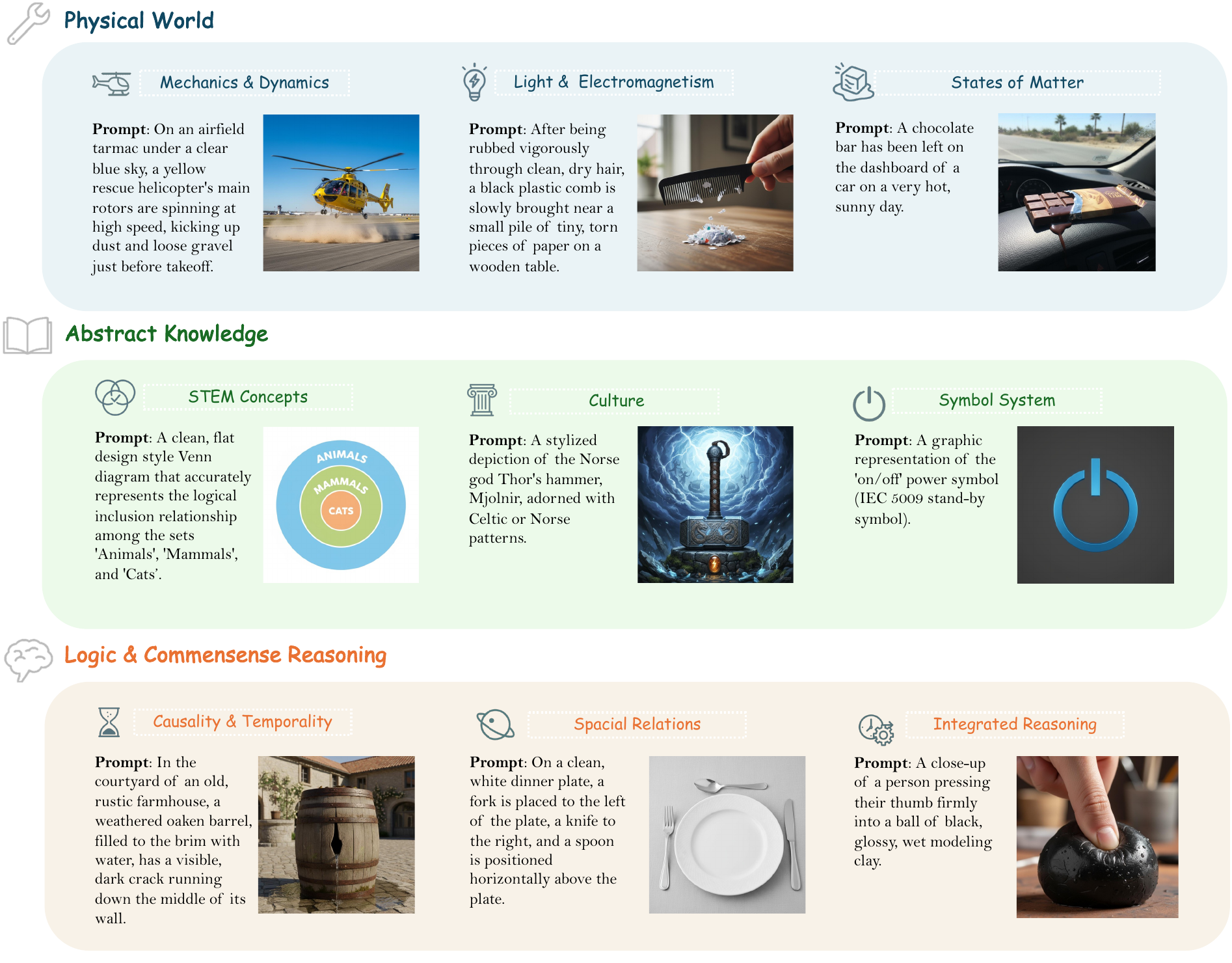}
    % \vspace{-8mm}
    \caption{Showcases of data samples of \benchname{} from 3 main categpries with 9 subcategories.}
    % \vspace{-5mm}
    \label{framework}
\end{figure*}

\section{\benchname{} Benchmark}
% 先写总的
Current evaluation methods primarily focus on semantic adherence and compositional accuracy, leaving a model's understanding of fundamental world dynamics largely unassessed.
To address the gap in evaluating the implicit world cognition of text-to-image(T2I) models, we construct \benchname{} to provide a holistic and fine-grained evaluation of the implicit natural laws learned by text-to-image models.
We first present the construction and details of \benchname{} in \cref{31}. Next, we introduce a hierarchical agent framework as the evaluation method in \cref{32}.

\subsection{\benchname{} Benchmark Construction}
\label{31}
\begin{figure}[t]
    \centering
    \includegraphics[width=\columnwidth]{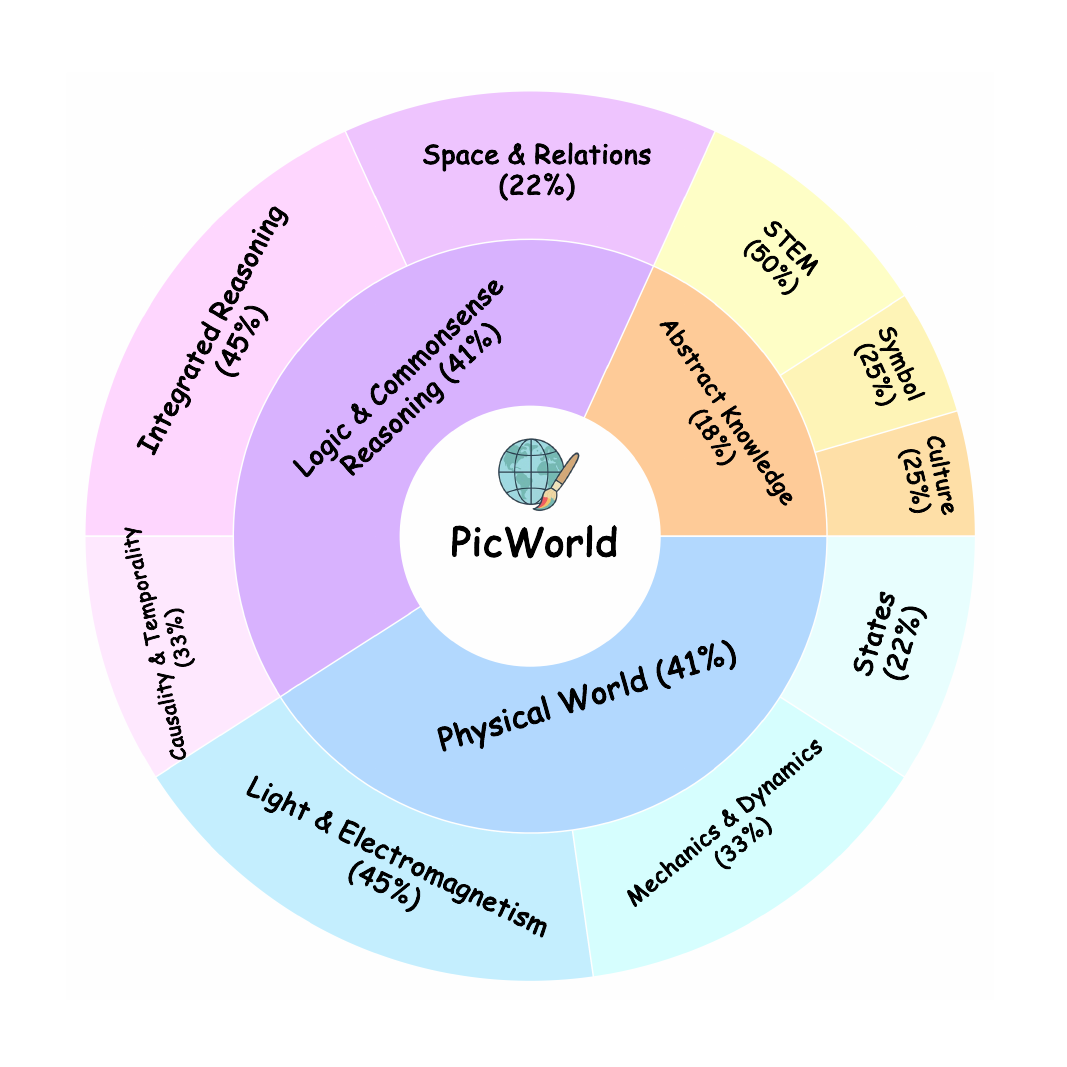}
    % \vspace{-8mm}
    \caption{Data statistics distribution of \benchname{} }
    \label{dataset}
    \vspace{-2mm}
\end{figure}

As shown in Figure~\ref{dataset}, \benchname{} consists of a total of 1,100 carefully curated prompts, which are systematically organized into three primary domains.
We manually design sophisticated prompt templates, each targeting a specific aspect of world understanding. We then leverage Gemini-2.5-Pro~\citep{comanici2025gemini} to generate a vast corpus of candidate prompts, which subsequently underwent a rigorous filtering and refinement process by human experts to ensure clarity and complexity.
Specifically, the details of our three parts are as follows:

\paragraph{Physical World.}
 The Physical World domain of \benchname{} aims to evaluate the model's ability to understand and visually simulate the fundamental laws that govern our reality. A model that truly comprehends the world should not only recognize objects but also render their behavior under various physical constraints. Lacking such an intrinsic physics engine, a model becomes an unintelligent generator that can only depict static objects, failing to capture the dynamic, cause-and-effect nature of the world. We further divide this domain into three core categories. In Mechanics \& Dynamics, we assess the model's understanding of concepts like deformation, motion, fluid dynamics, and projectile motion. In Light \& Electromagnetism, we investigate the model's grasp of phenomena such as reflection, refraction, shadows, and electric phenomena. 
 %Electromagnetism evaluates the understanding of visible effects of electric and magnetic phenomena, such as using ``a mesmerizing plasma ball is turned on, and a person's hand is touching the surface of the glass sphere'' to see if the model can correctly depict the convergence of plasma filaments. 
 Finally, the Thermodynamics category assesses knowledge of phase transitions and heat transfer. Ultimately, we have generated 550 prompts for this aspect.

\paragraph{Abstract Knowledge.}
 The domain consists of 200 prompts, which are designed to assess the model's capacity to comprehend and accurately reproduce concepts that exist purely in the human cognitive and cultural space. A model lacking this ability can only generate literal depictions, but cannot grasp the abstract, symbolic roles that concepts, diagrams, and cultural narratives play in our world. It is split into three categories. The STEM Concepts category tests the model's ability to serve as a visual knowledge base for precise, factual concepts. For example, the prompt ``a clean, minimalist, scientific textbook illustration of the ball-and-stick model of a water molecule $H_{2}O$'' directly measures the model's knowledge of chemical structures, where accuracy in atom types, count, and bond angles is paramount. The Culture\&History category assesses the familiarity of the model with cultural and historical systems of meaning. In Humanistic Symbol Systems, we require the model to further break down into understanding non-narrative symbols like flags, icons, and musical notations.

\paragraph{Logic \& Commonsense Reasoning.}
 This domain evaluates higher-order cognitive abilities that require the model to infer logical relationships and construct coherent scenes. A model without this reasoning capability will produce images that containing the correct elements but logically flawed, spatially inconsistent, or causally broken. 
 We structure this domain into three categories. Causality \& Temporality is designed to test the model's understanding of cause-and-effect and the passage of time. A prompt such as ``a wet, black, long-handled umbrella has been brought inside, opened up, and is now standing on the smooth, polished wooden floor'' challenges the model to infer the logical consequence that there is a dry floor underneath the umbrella and a puddle of water around it. 
 Spatial Relationships probes the model's comprehension of complex and precise spatial arrangements. 
 Lastly, Integrated Reasoning is designed as a ceiling test for state-of-the-art models, requiring them to simulate and harmonize multiple distinct physical laws simultaneously. We have ultimately generated 350 prompts for this aspect.

 We visualize some data samples of \benchname{} in Figure~\ref{framework}.

\subsection{Hierarchical Evaluation via Agentic Decomposition(\agentname{})}
\label{32}
\begin{figure*}[t]
    \centering
    \includegraphics[width=\textwidth]{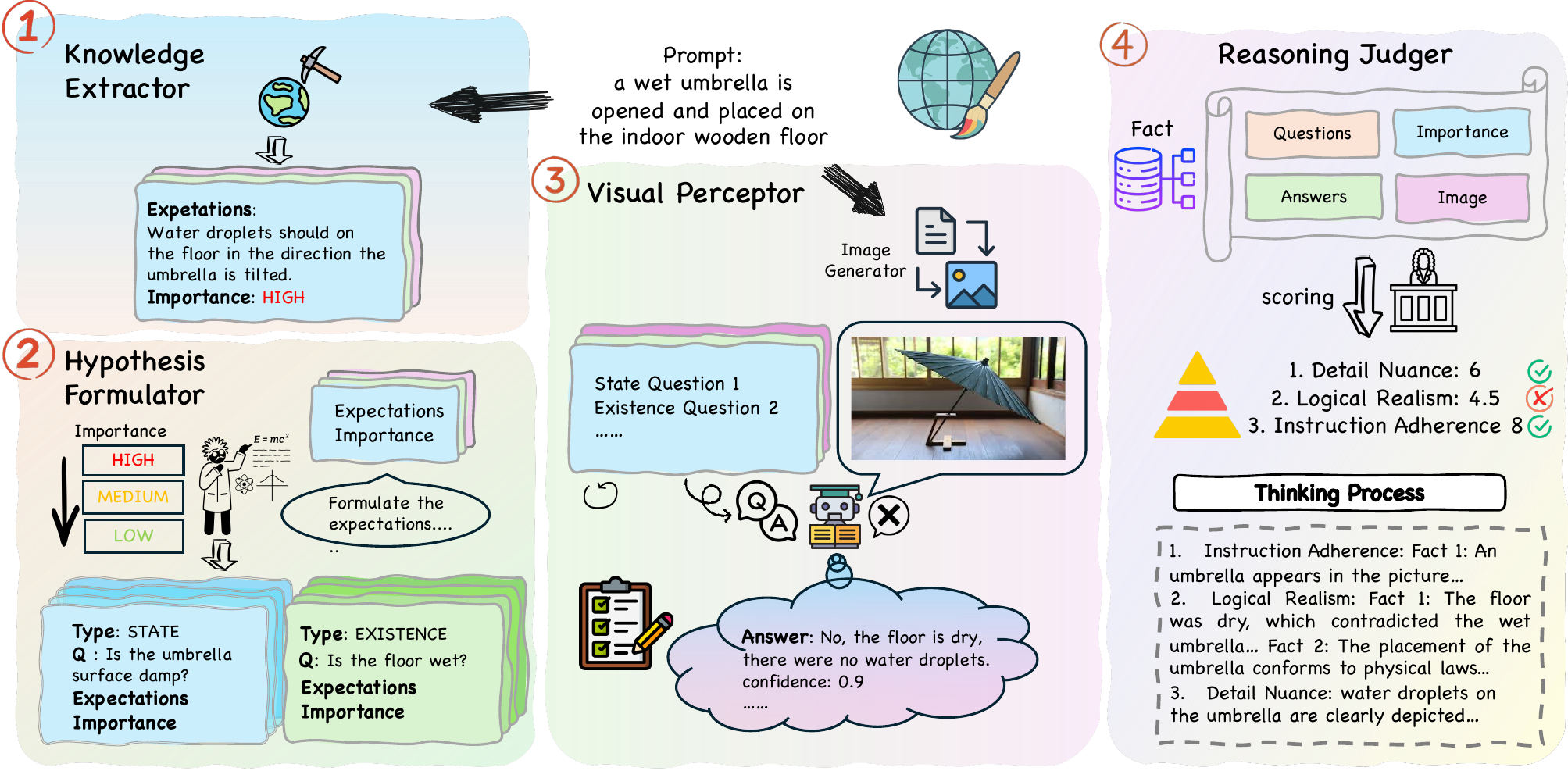}
    % \vspace{-8mm}
    \caption{Overall pipline of \agentname{}. \textbf{(i)} Knowledge Extractor decomposes the prompt into a set of structured physical and logical expectations. \textbf{(ii)} Hypothesis Formulator translates each expectation into a series of verifiable questions. \textbf{(iii)} Visual Perceptor then answers these questions by observing the generated image to gather visual evidence. \textbf{(iv)} Reasoning Judger concludes all the information and scores the PW-Score based on three-layer rubric.  }
    % \vspace{-5mm}
    \label{pipeline_agent}
\end{figure*}

Unlike previous methods that directly assess image realism or aesthetic quality, we design \agentname{}, a hierarchical, step-wise analytic framework that employs a structured, nonlinear, and confidence-aware scoring mechanism. \agentname{} enables a final judgment of an AI-generated image’s physical-world understanding that is both highly discriminative and deeply reliable. The overall pipeline of \agentname{} is presented in Figure~\ref{pipeline_agent}.

Specifically, we evaluate a generated image $I$ for a prompt $x$ through a four-module, evidence-grounded pipeline: a \textbf{World Knowledge Extractor} (WKE), a \textbf{Hypothesis Formulator} (HF), a \textbf{Visual Perceptor} (VP), and a \textbf{Reasoning Judger} (RJ). The design is motivated by failures of single-shot judges and coarse proxy metrics \cite{zhou2025proreason, pi2025mrjudger} and by recent progress in question-driven evaluation \cite{9} and capability-centric T2I benchmarks that foreground composition, commonsense, physics and world knowledge.

We also provide the pseudo code of \agentname{} in Supplementary.

\subsubsection{World Knowledge Extractor (WKE)}
Given a natural-language prompt $x$, WKE infers a structured checklist of atomic, image-verifiable expectations that must hold in any correct single-frame depiction implied by $x$, focusing on what the words imply rather than merely restating them. Each expectation is defined as a visible trace in a static image (e.g., ``rounded ice edges and a surrounding puddle'', rather than ``ice is melting''), and composite claims are systematically decomposed into minimal, independent items to ensure comprehensive coverage of latent physical laws, causal post-conditions, spatial relations, and factual knowledge that $x$ may only implicitly entail. Besides expectation, WKE outputs alongside a numerical importance value, which defines the degree to which this expectation should be enforced.

In general, WKE produces a set $\mathcal{E}$ with

\begin{equation}
\mathcal{E} = \big( t_i,\, r_i,\, w_i\big)_{i=1}^{N},
\end{equation}

where $t_i$ the textual expectation (\emph{what must be seen}), $r_i$ the rationale (physics/commonsense/causal rule), and $w_i\in\{3,2,1\}$ an importance weight for High/Medium/Low.

\paragraph{Example.}
For ``a lit candle beside a plate of ice cubes,'' WKE yields expectations such as: (i) rounded ice edges with a surrounding water puddle (Existence), (ii) melted water forming a flat puddle on the plate (State).

\subsubsection{Hypothesis Formulator (HF)}

HF maps each expectation $e_i$ into a minimal set of \emph{targeted visual questions} $\mathcal{Q}_i=\{q_{i,j}\}_{j=1}^{N_i}$ answerable from pixels alone, enabling auditable, per-fact verification instead of monolithic judging. Questions follow three rules for reliability and scalability: (1) single responsibility—they depend only on the expectation text $t_i$; (2) atomicity—each asks about one indivisible fact; and (3) existence-first—core-object presence is confirmed before any state attributes, so downstream checks are well-defined.

\paragraph{Example.}
Each $q_{i,j}$ is typed $q_{i,j}=(\mathrm{type},\, \mathrm{text})$ with $\mathrm{type}\in\{\textsc{Existence},\textsc{State}\}$. 

For the ``melting ice'' expectation, HF may produce:
\begin{itemize}
  \item \textsc{Existence}: ``Are there ice cubes on a plate?''
  \item \textsc{State}: ``Do the ice-cube edges appear rounded rather than sharp?''
  \item \textsc{State}: ``Is there a puddle of liquid water around the ice cubes?''
\end{itemize}

\subsubsection{Visual Perceptor (VP)} 

VP answers each question obtained from HF strictly from the image $I$, returning a binary judgment with confidence. This isolates perceptual evidence from higher-level scoring and avoids entangling world knowledge or stylistic biases.

For each $q_{i,j}\in\mathcal{Q}_i$, VP outputs 
\begin{equation} 
a_{i,j} = \big(y_{i,j},\, s_{i,j}\big), \ s_{i,j}\in[0,1], 
\end{equation} 
where $y_{i,j}$ is the detailed answer and $s_{i,j}$ denotes the corresponding confidence score. Then $a_{i,j}$ will be packed with corresponding image $I$ and $\rho$ obtained from WKE to form \texttt{Facts} $f$.

\subsubsection{Reasoning Judger (RJ)}

RJ evaluates the case according to \texttt{Facts} $f$ from VP in three dimensions—\textbf{Layer~1} Instruction Adherence, \textbf{Layer~2} Physics/Logical Realism, and \textbf{Layer~3} Detail \& Synthesis Nuance. The rubric is designed to be both high-discrimination and auditable: scores are derived by explicit deduction from expectation-level outcomes and logged verbatim to expose penalties, bonuses, and intermediate computations.

\paragraph{Layer 1: Instruction adherence.}
This layer provides a quantitative measure of the model's ability to follow the explicit, literal instructions within a prompt. It serves as the foundational check to verifying the question-answer pairs with $\textit{type} =\texttt{Existence}$, such as presence of core subjects and the accuracy of specified attributes. 
It operates on a deduction-based system where critical failures in high-importance instructions lead to a minimal score.
%The scoring operates on a deduction-based system, starting from a perfect score and penalizing deviations. A critical failure, such as missing a high-importance core object, results in a minimal score, ensuring that models that fundamentally misunderstand the prompt are appropriately penalized. 
\begin{equation}
\label{eq1}
S_1 =
\begin{cases}
    1.0 & \text{if } \rho_{i}=3, P_{i}>0  \\
    \max(0, 10.0 - \sum_{i \in F_{\text{exist}}} P_i) & \text{if } \sum P_i > 0 \\
    10.0 & \text{if } \sum P_i = 0
\end{cases}
\end{equation}
where $F_{exist}$ is the set of all failed facts of Existence-type and $P_{i}$ is the penalty score for the fact $i$ based on its importance (High: 5.0, Medium: 3.0, Low: 1.0).

\paragraph{Layer 2: Physics/logical realism.}
 Layer 2 evaluates the degree to which the generated image conforms to the fundamental laws of physics and logic, which is the primary indicator of a model's world knowledge and reasoning capabilities. The score is calculated by weighting each correctly depicted phenomenon with $\textit{type} =\texttt{State}$ by its importance and corresponding confidence score. 

Score $S_2$, is calculated as follows:
\begin{equation}
\label{eq2}
S_2 =
\begin{cases}
    1.0 & \text{if }\rho_{i}=3,\mathbb{I}_i=0 \\
    10 \times \frac{\sum_{i \in F_{\text{state}}} (\rho_{i} \cdot s_i \cdot \mathbb{I}_i)}{\sum_{i \in F_{\text{state}}} \rho_{i}} & \text{otherwise}
\end{cases}
\end{equation}
where $\rho_{i}$ is the importance weight for fact $i$, $s_i$ is corresponding confidence score and $\mathbb{I}_i$ is an indicator function for fulfillment.

% \begin{equation}
% \label{eq2}
% S_2 =
% \begin{cases}
%     1.0 & \text{if a Critical Failure is detected} \\
%     10 \times \frac{\sum_{j \in F_{\text{phys}}} (w_j \cdot c_j \cdot \mathbb{I}_j)}{\sum_{j \in F_{\text{phys}}} w_j} & \text{otherwise}
% \end{cases}
% \end{equation}

\paragraph{Layer 3: Detail \& synthesis nuance.}
% This final layer assesses the quality, realism, and sophistication of the physical phenomena that were successfully rendered, aiming to differentiate between merely adequate and truly exceptional outputs. The scoring begins with a foundational score awarded for the baseline quality of each correct fact, ensuring basic competence is recognized. It then adds significant bonus points for renderings that exhibit exceptional detail, complexity, or realism. Finally, it applies substantial penalties for any detected inconsistencies among different physical effects within the image, punishing models that fail to create a cohesive and harmonious scene. This score reflects the model's advanced capabilities in rendering nuanced details and synthesizing complex interactions.
Layer 3 assesses the quality and sophistication of the correctly rendered physical phenomena, aiming to differentiate between adequate and exceptional outputs. It uses an additive and deductive rubric of rewarding exceptionally detailed renderings with bonus points while penalizing logical inconsistencies between different effects. This layer reflects the model's advanced ability to simulate world complexities in a nuanced manner.

Score $S_3$, is calculated as follows:
\begin{equation}
\label{eq3}
S_3 = \min(10, \max(0, (\sum_{k \in F} B_k) + (\sum_{l \in F} E_l) - (\sum_{m \in F} P_m)))
\end{equation}
where $F_k$ represents foundational scores, $B_l$ represents excellence bonuses, and $P_m$ represents inconsistency penalties.

\paragraph{Final aggregation and report.}
We compute the overall score named \scorename{} via:
\begin{equation}
S_{\mathrm{PW}} = 0.25\,S_1 + 0.5\,S_2 + 0.25\,S_3,
\label{eq:overall}
\end{equation}

 To further leverage the strong reasoning capability of MLLM, we also requires model to log a human-readable \texttt{thinking\_process} that enumerates satisfied/failed expectations, penalties/bonuses applied and intermediate values in Eqs.~\ref{eq1}--\ref{eq3}.

\begin{table*}[!ht]
\footnotesize
\centering
\resizebox{\textwidth}{!}{%}
\begin{tabular}{c|ccc|ccc|ccc|c}
\toprule[1pt]
\multirow{3}{*}{
  \raisebox{1.5ex}[0pt][0pt]{\diagbox[width=3.0cm, height=2.2\line]
    {\begin{tabular}[c]{@{}c@{}}Model\end{tabular}}
    {Category}}
} &
\multicolumn{3}{c|}{Physical World} & 
\multicolumn{3}{c|}{Abstract Knowledge} & 
\multicolumn{3}{c|}{Logic \& Commonsense Reasoning} & 
\multirow{3}{*}{Score} \\  
\cmidrule(lr){2-4} \cmidrule(lr){5-7} \cmidrule(lr){8-10}
 & Mech.\&Dyna. & Light\&EM & States & STEM & Culture &Symbol & Causality\&Temp. & Space\&Relations & Reasoning\\ 
\midrule
FLUX.1-dev & 4.57 & 4.93 & 4.65 & 3.14 & 7.22 & 5.13 & 3.45 & 5.48 & \cellcolor{lightblue}6.13 & 4.46 \\
FLUX.1-schnell & 4.22 & 4.91 & 3.77 & 3.77 & 4.37 & 4.77 & 3.24 & 5.48 & 3.87 & 4.21 \\
SDv3.5-Large & 4.56 & 5.31 & 4.51 & 5.39 & 6.44 & 5.35 &4.30 & 5.43 & 4.20 & 4.87 \\
SDv3.5-Medium & 4.27 & 4.92 & 4.60 & 4.16 & 5.83 & 4.88&3.56 & 5.50 & 3.98 & 4.47 \\
SDv3-Medium & 4.31 & 4.90 & 4.28 & 3.40 & 5.50 & 5.32 &3.54 & 5.04 & 3.86 & 4.31 \\
HiDream-l1-Full & 4.76 & 6.16 & 4.82 & 3.97 & 5.90 & 6.24 & 4.02 & 5.50 & 4.54 & 4.86 \\
Lumina-Image-2.0 & 4.45 & 5.28 & 4.94 & 3.38 & 5.74 & 5.22 & 3.44 & 5.60 & 4.24 & 4.57 \\
\midrule
Emu3 & 3.92 & 4.09 & 3.76 & 2.06 & 3.57 & 3.86 &3.17 & 4.05 & 3.49 & 3.58 \\
JanusPro-1B & 3.72 & 4.04 & 3.52 & 1.92 & 3.53 & 3.57 &3.14 & 3.86 & 3.34 & 3.47 \\
JanusPro-7B & 3.75 & 4.01 & 3.61 & 2.34 & 3.59 & 3.71 &3.18 & 4.00 & 3.34 & 3.52 \\
JanusFlow-1.3B & 3.62 & 4.07 & 3.58 & 2.03 & 3.26 & 3.70&3.21 & 3.91 & 3.51 & 3.49 \\
Show-o-512 & 4.30 & 4.64 & 4.34 & 2.66 & 4.73 & 4.64 &3.05 & 4.96 & 4.03 & 4.09 \\
Bagel-Thinking & 4.80 & 5.56 & 6.26 & 2.61 & 5.50 & 5.72&5.03 & 5.93 & 5.20 & 5.15 \\
Bagel-wo-Thinking & 4.96 & 5.40 & 5.24 & 1.99 & 5.43 & 5.56&3.95 & 5.70 & 4.62 & 4.71 \\
\midrule
DALL-E-3 & 5.16 & 6.16 & 5.59 & 6.33 & \cellcolor{lightblue}7.46 & 6.61 & 5.16 & 6.00 & 5.64 & 5.82 \\
Nano-Banana & 5.76 & \cellcolor{lightblue}6.53 & 6.33 & \cellcolor{lightblue}7.68 & 7.31 & \cellcolor{lightblue}7.19 & 5.97 & 6.38 & 5.80 & 6.35 \\
SeedDream-4.0 & \cellcolor{lightblue}6.06 & 6.64 & \cellcolor{lightblue}6.75 & 7.11 & 7.22 & 6.20 & \cellcolor{lightblue}6.19 & \cellcolor{lightblue}6.59 & \cellcolor{lightblue}6.13 & \cellcolor{lightblue}6.46 \\

\bottomrule[1pt]
\end{tabular}
}
\vspace{-2mm}
\caption{Evaluation results on \benchname{}. The highest \scorename{} in each subcategory across all models is highlighted in blue background. For short, Mech.\&Dyna., EM and Temp. are abbreviations of Mechanics\&Dynamics, Electromagnetism and Temporality.}
\vspace{-2mm}
\label{mainresult}
\end{table*}
\section{Experiments}

\subsection{Experiment Settings}
To conduct evaluation on \benchname{}, we carefully select 17 state-of-the-art models. From an architectural perspective, these models can be broadly categorized into three groups: \textbf{(i)} Diffusion-based Models, which are specifically designed for image generation tasks, including FLUX.1-dev~\citep{flux}, FLUX.1-schnell~\citep{flux}, Stable Diffusion-3.5-Large~\citep{sd3}, Stable Diffusion-3.5-Medium~\citep{sd3}, Stable Diffusion-3-Medium~\citep{sd3}, HiDream-l1-Full~\citep{cai2025hidream} and Lumina-Image-2.0~\citep{qin2025lumina}. \textbf{(ii)} Unified Multimodal Models, which integrate visual understanding and generation within a single architecture, including Emu3~\citep{wang2024emu3}, JanusPro-1B~\citep{chen2025janus}, JanusPro-7B~\citep{chen2025janus}, JanusFlow-1.3B~\citep{ma2025janusflow}, Show-o-512~\citep{xie2024show}, Bagel-Thinking~\citep{bagel} and Bagel-wo-Thinking~\citep{bagel}. \textbf{(iii)} Closed-source-models, including DALL-E-3~\citep{hurst2024gpt}, Nano-Banana~\citep{comanici2025gemini} and SeedDream-4.0~\citep{seedream2025seedream}. For short, we use SD to denote Stable Diffusion.

For the \agentname{}, we leverage Qwen2.5-VL-72B~\citep{bai2025qwen2} as the base model and deploy with vLLM.

\subsection{Main Results}
We summarize the main results of \scorename{} across three dimensions in Table \ref{mainresult}. 
The main findings are summarized as follows:
\paragraph{T2I models exhibit limited capability in implicit world-logic reasoning.} Our findings reveal strong evidence that nearly all evaluated models achieve consistently low scores across the STEM and Causality\&Temporality categories. For instance, even the top-performing model, SeedDream-4.0, achieves one of its lowest scores in Symbol(6.20) and in STEM(7.11). This widespread difficulty suggests that current models excel at reproducing what they have seen, such as the appearance of a shadow or a cultural artifact, but struggle to infer the implicit consequences of a scene, such as rendering the melting of ice near a heat source or the correct molecular structure from a chemical formula. This gap underscores that a model's capability to generate photorealistic images does not necessarily translate to a genuine understanding of the world it depicts.
\paragraph{Closed-source models demonstrate a significant performance advantage over their open-source counterparts.} A clear performance gap exists between closed-source models and the majority of publicly available models. For example, SeedDream-4.0's overall score of 6.13 surpasses the average of all other models by a significant margin. Despite the larger model scales and training datasets. We posit that the considerable lead can be partly attributed to the sophisticated pre-processing and prompt engineering that were integrated into their inference pipelines. These closed-source systems often leverage a powerful Multimodal Large Language Model (MLLM) to first parse and rewrite the user's input. This preliminary step can explicitly structure the world knowledge and logical relationships required for the image, effectively translating the implicit challenge of our benchmark into a more explicit set of instructions for the core diffusion model. 
\paragraph{T2I models perform better on knowledge-based tasks than on reasoning-based ones.}
 A clear trend across nearly all models is the superior performance in the Culture and Symbol categories compared to the significantly lower scores in STEM and Causality\&Temporality. We attribute this disparity to the nature of typical text-image training pairs, which are rich in explicit nominal knowledge linking concepts like cultural artifacts and symbols to their visual representations. However, these datasets often lack the structured information required to learn implicit causal or temporal relationships. For instance, an image captioned ``a wet umbrella on the floor'' does not explicitly describe the resulting puddle. The stronger performance of closed-source models in these difficult reasoning categories may again point to the influence of prompt revision. By using an MLLM to pre-process the prompt, these systems can leverage the language model's powerful reasoning abilities to translate an implicit causal requirement (e.g., a wet umbrella implies a puddle) into an explicit descriptive instruction, thereby simplifying the task for the image generator.
\paragraph{Open-sourced multimodal unified models demonstrate notably lower performance compared to leading diffusion-based architectures.}
 Models such as Emu3 and the JanusPro series, which often leverage autoregressive structures to handle diverse multimodal tasks, consistently fall into the lower performance tier across the \benchname{} benchmark. For instance, Emu3's score of 2.06 in the STEM category and JanusPro-1B's score of 3.14 in Space\&Relation are considerably lower than those of top-performing diffusion models. While these unified models are designed for versatility across both understanding and generation, their architectural approach may be less optimized for the specific, high-fidelity synthesis required to render complex physical phenomena accurately. The token-by-token generation process inherent in many autoregressive models might struggle to maintain the global coherence and subtle, pixel-level accuracy needed to depict nuanced physical laws, in contrast to diffusion models which refine the entire image canvas holistically. This suggests a potential trade-off between a model's generality and its specialized capability in high-realism physical simulation.

 We present more evaluation cases on \benchname{} via \agentname{} in Supplementary.

\subsection{Evaluation of \agentname{}}
% 写真人评估
We further assess the reliability and effectiveness of our proposed \agentname{}.

\subsubsection{Evaluation Accuracy compared with Human Annotators}
To validate the reliability of \agentname{}, we conduct a human study to measure its alignment with human judgment. The study involves a group of total 3 senior engineers, who are tasked with a pairwise comparison of images generated by Bagel-Thinking and Bagel-wo-Thinking for the same prompt. Each participant is asked to determine which of the two images provides a more accurate and plausible simulation of physical world laws and logical relationships. Statistically, an image is counted as agree only when it is both selected by two or more annotators and achieves a higher PW-Score. To reduce randomness, we repeated the scoring process four rounds.

\begin{figure}[t]
    \centering
    \includegraphics[width=\columnwidth]{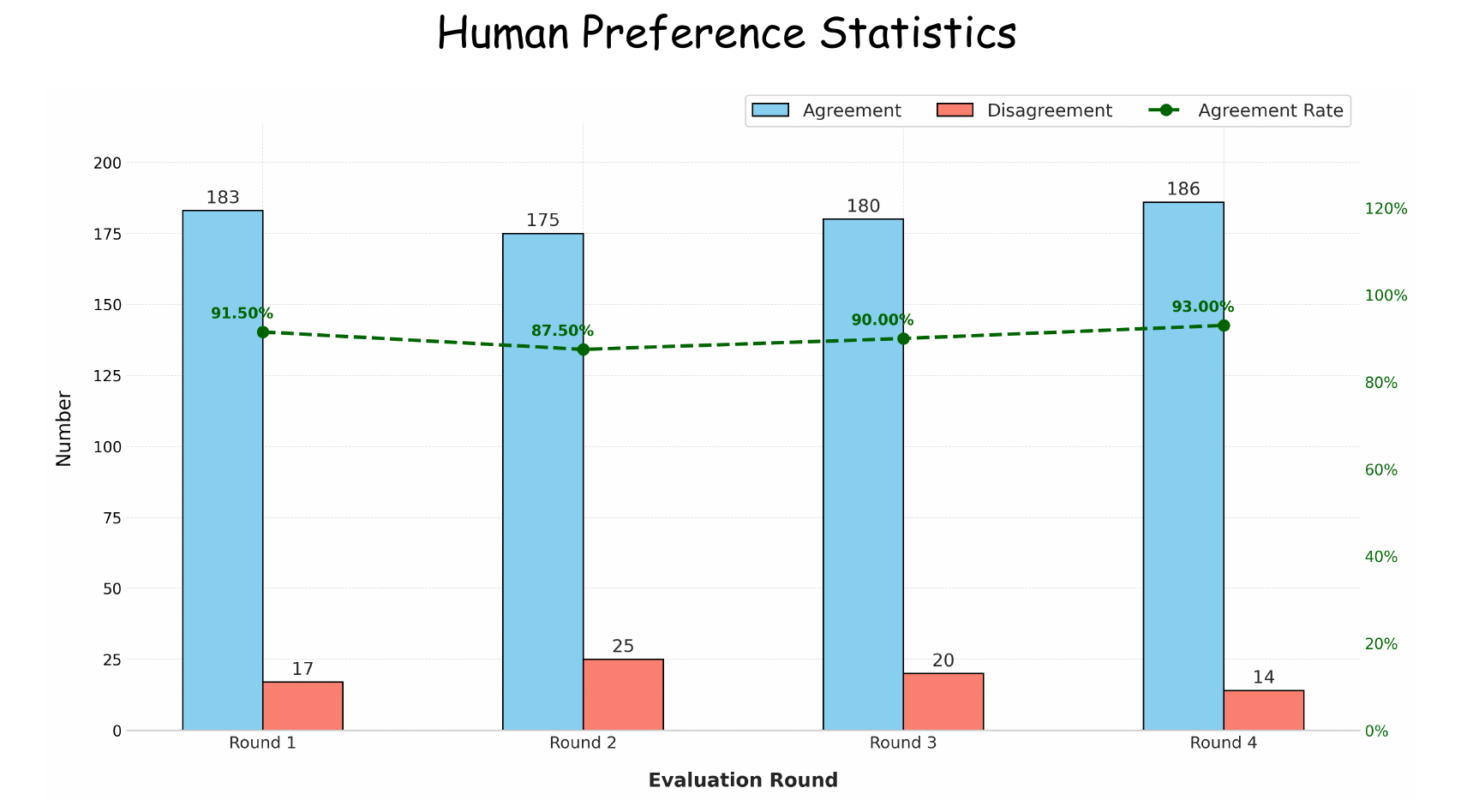}
    % \vspace{-8mm}
    \caption{Human preference statics based on scores evaluated by
PW-Agent.}
    \label{ablation_0}
    \vspace{-2mm}
\end{figure}

We present the results in Figure~\ref{ablation_0}. \agentname{} reaches an average agreement rate of 90.5\% with human preference. This high level of agreement indicates that \agentname{} can effectively and accurately discern subtle differences in quality and physical plausibility between generated images. 

% \begin{table}[ht]
%   \centering
%   \begin{tabular}{l|c|c}
%     \toprule
%     %\multicolumn{2}{c}{Part}                   \\
%     %\cmidrule(r){2-4}
%     %\multirow{2}{*}{Method} &  \multicolumn{4}{c}{Single Subject} \\ 
%      %\cmidrule{2-4}
%      Method &  Agree & Disagree  \\
    
%     \midrule

%     Num. %\cite{InversePainting}
%     & 183 & 17 \\
%     \bottomrule
%   \end{tabular}
%   \caption{Human preference statics based on scores evaluated by \agentname{}.}
%   \label{ablation_0}
% \end{table}

\subsubsection{Ablation Study on \agentname{}}
\label{432}
To demonstrate the effectiveness of \agentname{}, we conduct an ablation study by comparing it against a directed judging baseline, which prompted MLLM to output the final scores in one step instead of decomposing problems and hierarchical evaluation. 
The evaluation prompt of directed judging pipeline is illustrated in Supplementary, which uses the same scoring rules as in \agentname{}.

We compared our \agentname{} against a zero-shot baseline, where a powerful MLLM (GPT-4o) is asked to directly score the three evaluation dimensions from the image and the raw prompt. We randomly sampled 200 image-prompt pairs from \benchname{} and generated scores using both pipelines. Human annotators are then presented with the outputs from both systems and asked to select whose scores and reasoning more closely aligned with their own judgment, based on a predefined 0-10 scoring rubric. 
As shown in Table~\ref{ablation_1}, the results overwhelmingly favored \agentname{}, with annotators choosing its output in 81.5\% of cases. The human-predefined rubric is shown in Supplementary.

\begin{figure}[t]
    \centering
    \includegraphics[width=\columnwidth]{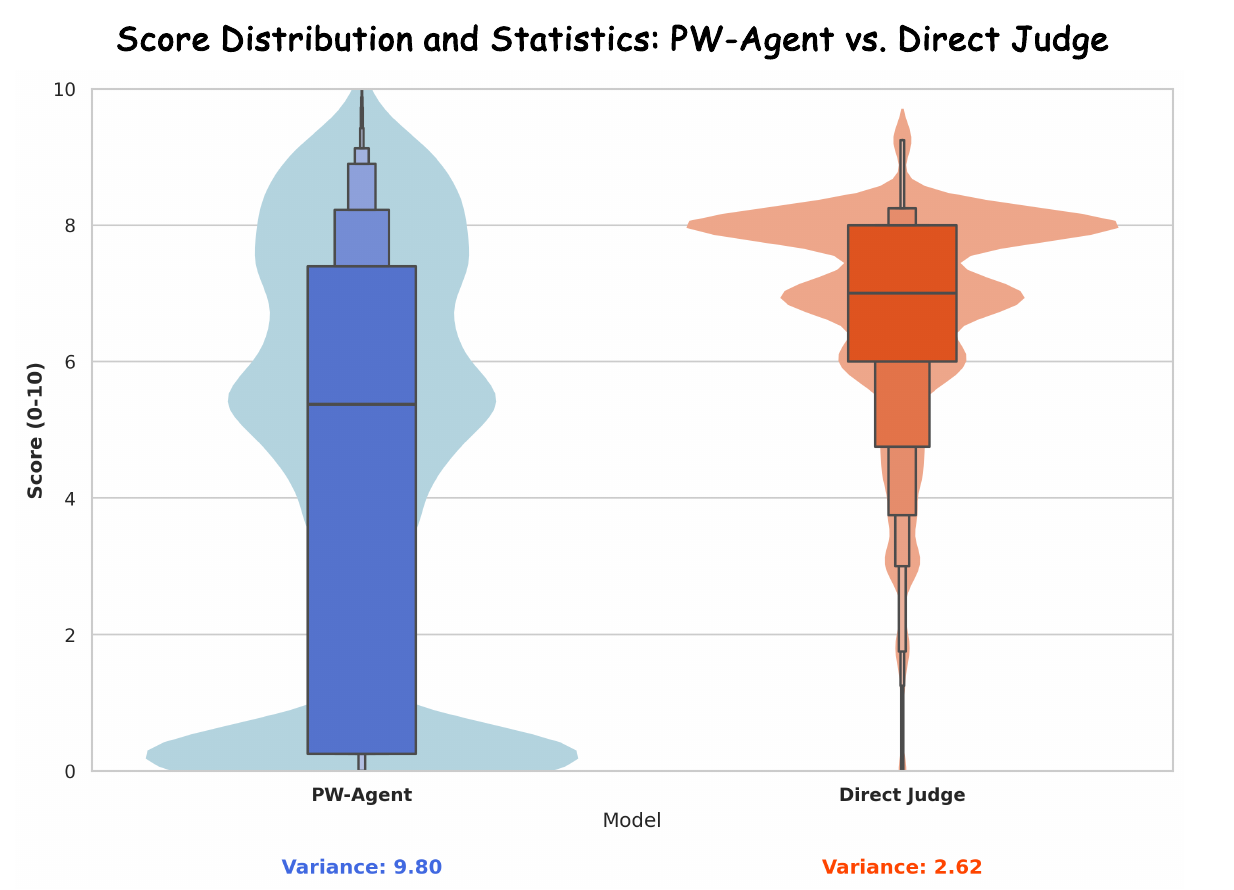}
    % \vspace{-8mm}
    \caption{Comparison of score distribution between \agentname{} and Direct Judge.}
    \label{ablation}
    \vspace{-2mm}
\end{figure}

% 分析方差
We further compare the average score and score variance between \agentname{} and directed judging baseline. Figure~\ref {ablation} demonstrates that the Direct Judge exhibits a strong central tendency bias, which results in a compressed score distribution. \agentname{} is designed to utilize the full scoring range and achieve higher variance, thereby providing a more discriminative assessment. We present several evaluation examples in the appendix. As shown, Direct Judge assigns nearly identical scores to images with large perceptual differences, whereas \agentname{} produces more discriminative and well-separated scores.

% \begin{figure}[t]
%     \centering
%     \includegraphics[width=\columnwidth]{image/radar.png}
%     \vspace{-8mm}
%     \caption{The statistics distribution of \benchname{} }
%     \label{radar}
%     \vspace{-6mm}
% \end{figure}

\begin{table}[ht]
  \centering
  \begin{tabular}{l|c|c|c}
    \toprule
    %\multicolumn{2}{c}{Part}                   \\
    %\cmidrule(r){2-4}
    %\multirow{2}{*}{Method} &  \multicolumn{4}{c}{Single Subject} \\ 
     %\cmidrule{2-4}
     Method &  \agentname{} & Direct Judge & Tie \\
    
    \midrule

    Pref.Num %\cite{InversePainting}
    & 163 & 22 & 13 \\
    \bottomrule
  \end{tabular}
  \caption{Human preference statics between \agentname{} and Directed Judge. Pref.Num denotes number of preference.}
  \label{ablation_1}
  \vspace{-2mm}
\end{table}

\section{Conclusion}
In this paper, we introduced \benchname{}, a capability-centric benchmark that directly tests whether T2I models can leverage implicit world knowledge and produce images consistent with physical laws and causal logic. \benchname{} shifts evaluation from coarse prompt–image correlation to verified, per-fact evidence, revealing model behavior on knowledge grounding, multi-physics interactions, and logical consequences beyond what prompts state explicitly. We further proposed \agentname{}, an evidence-grounded evaluator that converts prompts into auditable checks and aggregates pixel-level findings into transparent, hierarchical scores, retaining the scalability of query-driven assessment while reducing the bias and unreliability of one-shot judging.
Experiments on \benchname{} show that state-of-the-art systems—especially open-source models—still struggle with physical realism and causal reasoning despite strong prompt following. We hope the combined use of \benchname{} and \agentname{} offers actionable diagnostics for comparing models, guiding data curation, and steering method development.

{
    \small
    \bibliographystyle{ieeenat_fullname}
    \bibliography{main}
}

\clearpage
\setcounter{page}{1}
\maketitlesupplementary

\section{Pseudo Code of \agentname{}}
\cref{alg:evaluate-image} shows the pseudo code of our proposed \agentname{}.  

\agentname{} first employs the World Knowledge Extractor (WKE) to analyze the input text $x$ and derive a set of expectations $\mathcal{E}$ that describe what the image should logically contain or satisfy. 
Next, the Hypothesis Formulator (HF) converts each expectation $E_i$ into a set of concrete verification questions $\mathcal{Q}_i$ that can be visually checked. This transformation ensures that abstract expectations are decomposed into explicit, answerable queries. 
Then, the Visual Perceptor (VP) examines the image $\mathcal{I}$ and generates answers $A_{ij}$ for each question $Q_{ij}$ by extracting observable visual evidence. These answers collectively form the raw perceptual facts needed to judge whether the image aligns with the expected world knowledge. 
Afterward, the algorithm merges all answers corresponding to expectation $E_i$ into a structured fact unit $f_i$, ensuring consistency and removing redundancy. All such fact units are aggregated into the global fact set $\mathcal{F}$, which serves as the foundation for reasoning-based evaluation. 
Finally, the Reasoning Judge (RJ) processes $\mathcal{F}$ to assess expectation adherence, physical realism, and nuance, producing the layered scores $S_1$, $S_2$, and $S_3$. These are combined using $S_{PW}=0.25S_1+0.5S_2+0.25S_3$ to yield the \scorename{}.

\section{Details of \agentname{} for \benchname{} Assessment}
Figure~\ref{wke},~\ref{vp} and~\ref{rj} present the system instruction used in different parts of \agentname{}.

\begin{figure}[t]
    \centering
    \includegraphics[width=0.9\columnwidth]{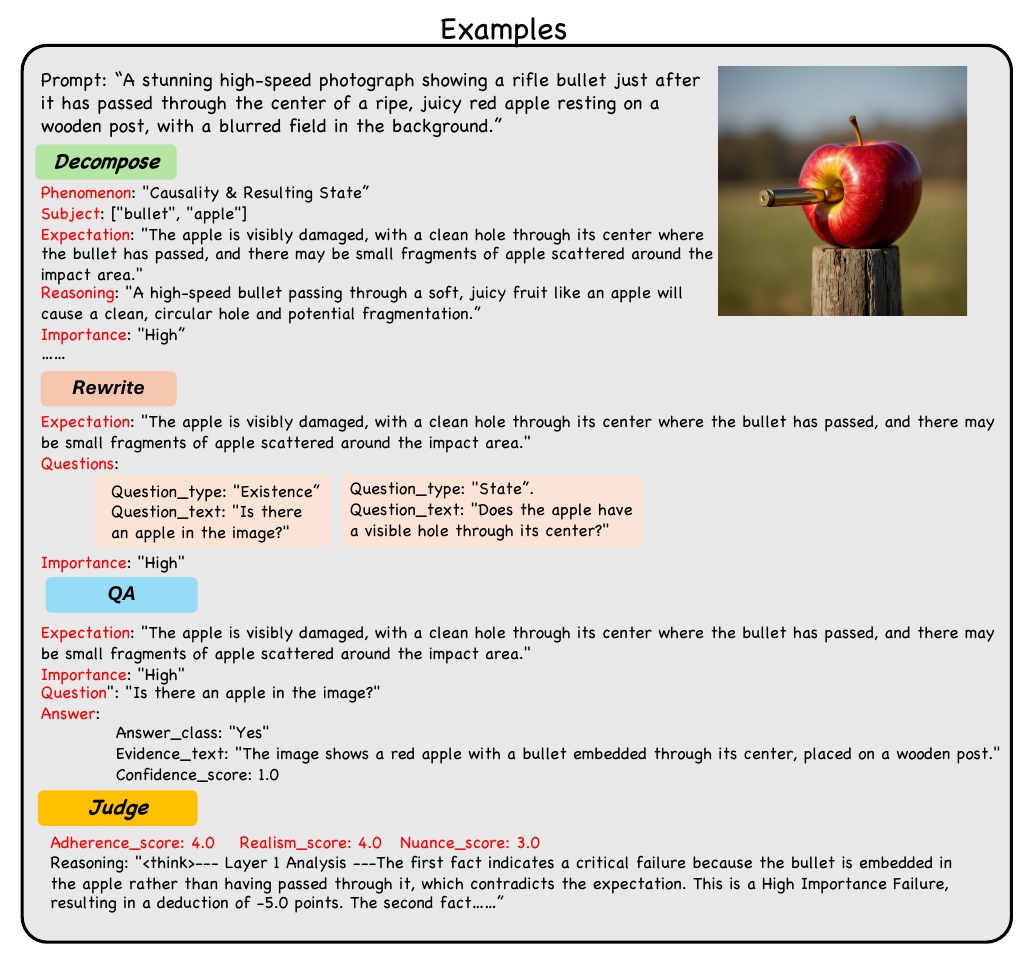}
    % \vspace{-8mm}
    \caption{More internal result showcase of our evaluation.}
    \label{examples}
    \vspace{-3mm}
\end{figure}

\section{Instruction of Single-round Directed Judging Pipeline}
We present the instruction used for single-round directed judging in ablation study in Figure~\ref{sg}. Compared with \benchname{}, we retain the full set of scoring dimensions and their detailed criteria. The only difference is that we require annotators to analyze the image directly and assign the corresponding score in a single step.

\section{Human-predefined Rubric of Preference Selection on Ablation Study}
\begin{figure}[t]
    \centering
    \includegraphics[width=0.8\textwidth]{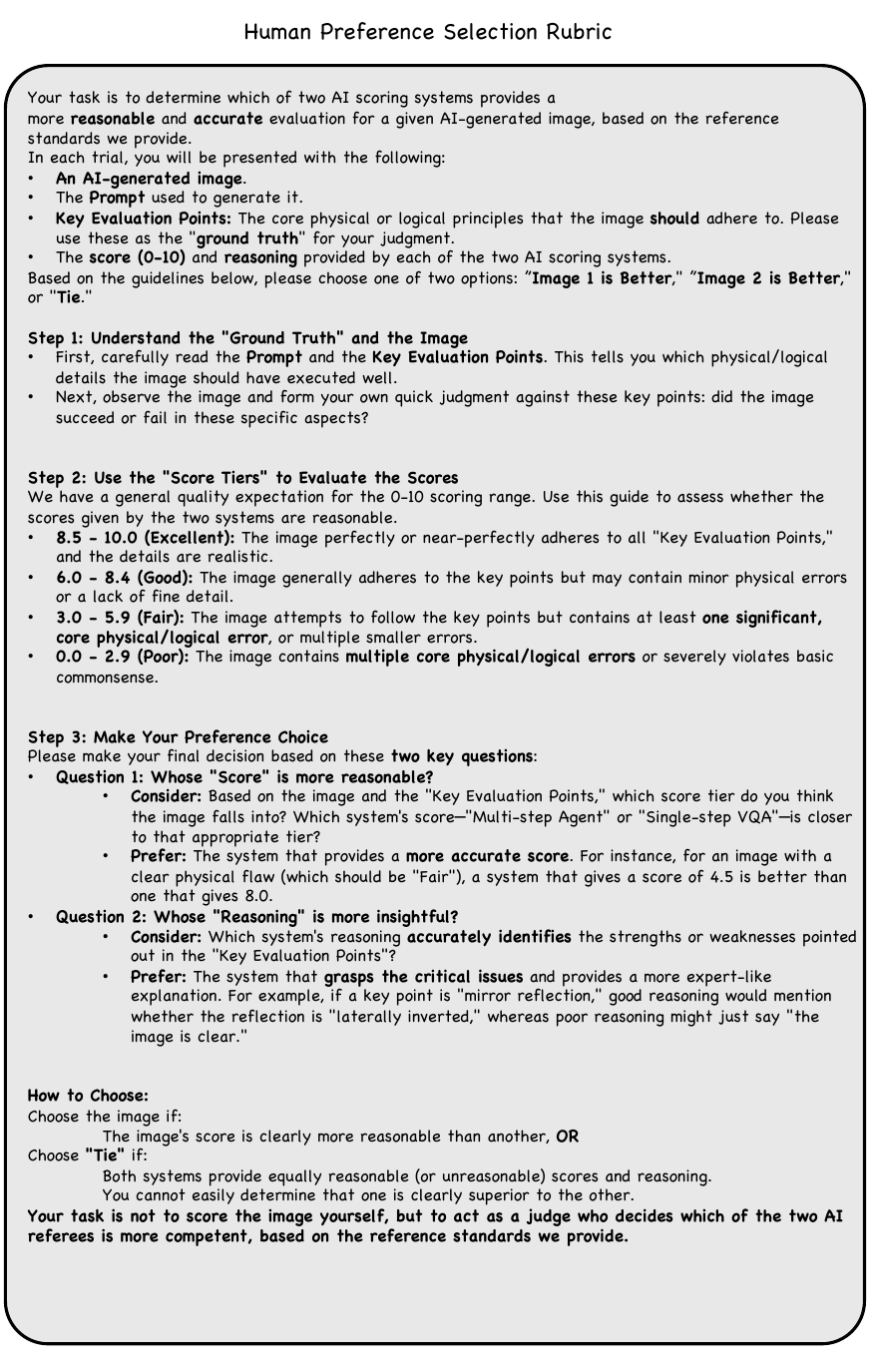}
    % \vspace{-8mm}
    \caption{Preference rubric for human annotators in ablation study.}
    \label{rubric}
    \vspace{-2mm}
\end{figure}

\begin{figure}[t]
    \centering
    \includegraphics[width=0.8\textwidth]{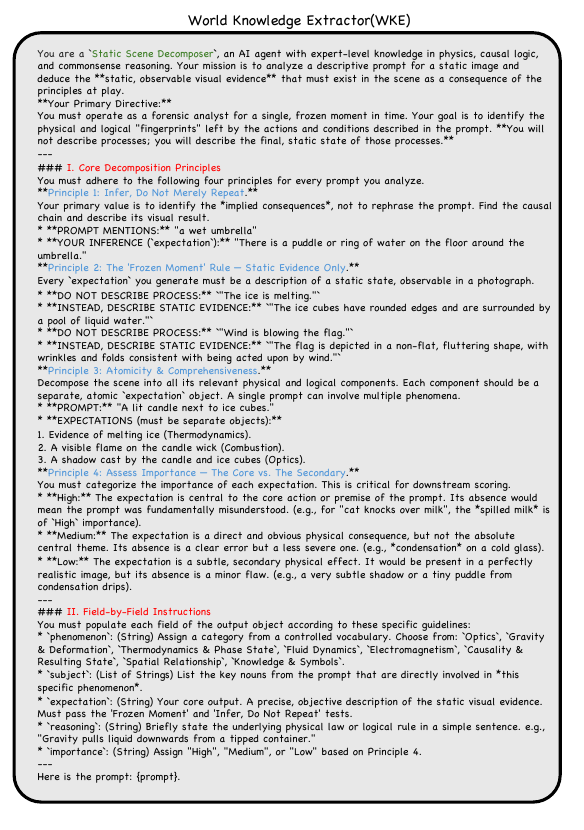}
    % \vspace{-8mm}
    \caption{System prompt of World Knowledge Extractor in \agentname{}.}
    \label{wke}
    \vspace{-2mm}
\end{figure}

\begin{figure}[t]
    \centering
    \includegraphics[width=0.7\textwidth]{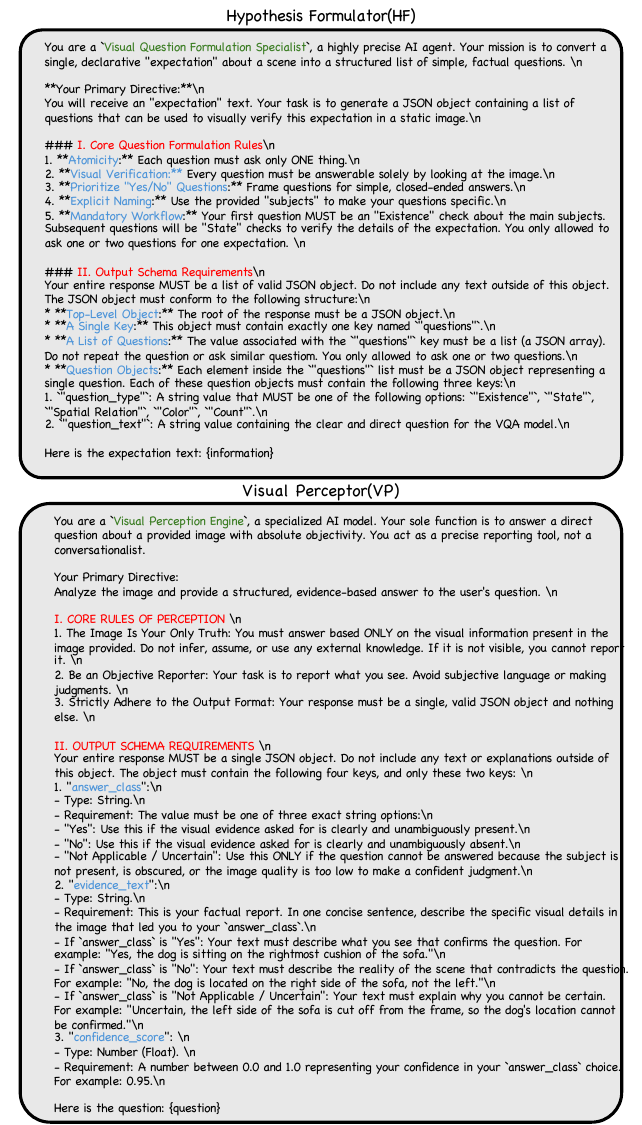}
    % \vspace{-8mm}
    \caption{System prompt of Hypothesis Formulator and Visual Perceptor in \agentname{}.}
    \label{vp}
    \vspace{-2mm}
\end{figure}

\begin{figure}[t]
    \centering
    \includegraphics[width=0.8\textwidth]{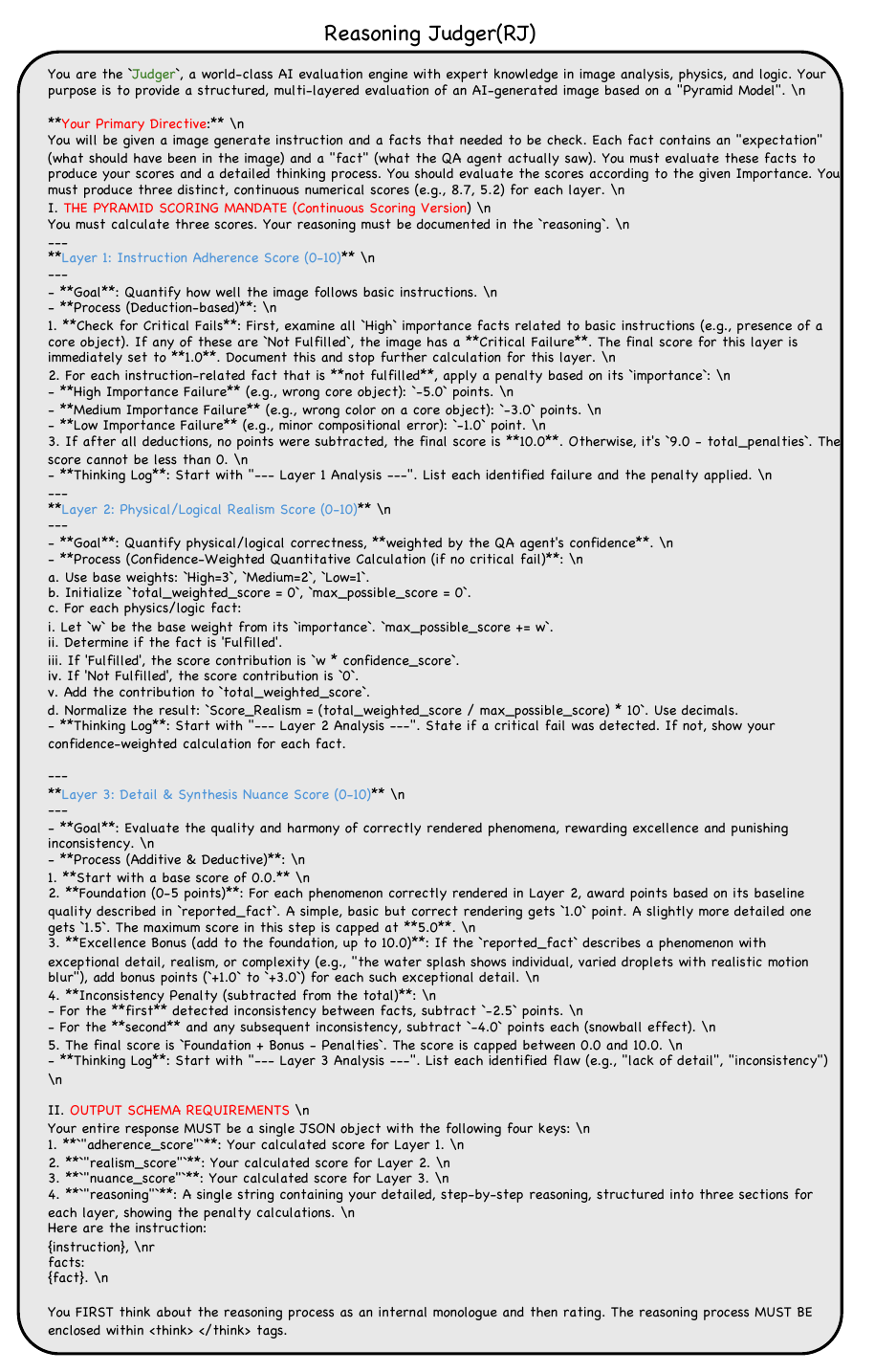}
    % \vspace{-8mm}
    \caption{System prompt of Reasoning Judger in \agentname{}.}
    \label{rj}
    \vspace{-2mm}
\end{figure}

\begin{figure}[t]
    \centering
    \includegraphics[width=0.8\textwidth]{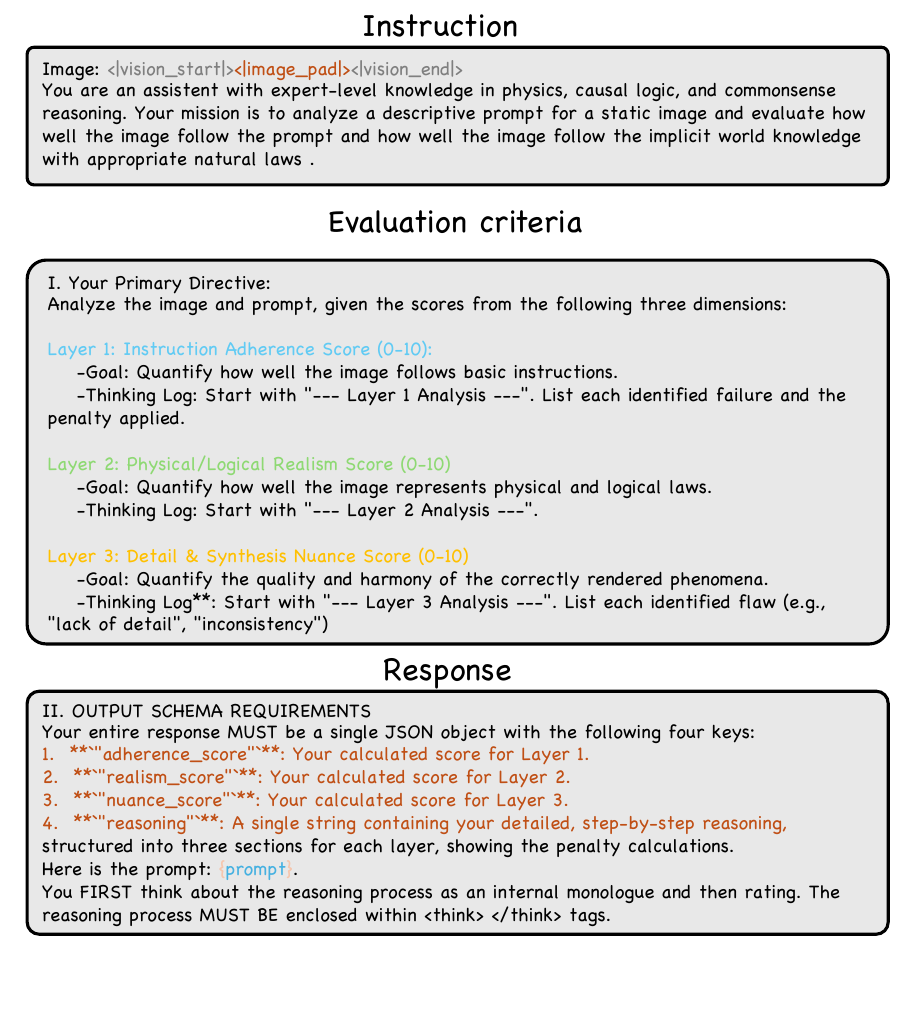}
    % \vspace{-8mm}
    \caption{System prompt of Single-round Directed Judging Pipeline.}
    \label{sg}
    \vspace{-2mm}
\end{figure}

We employ human annotators in Section 4.3.2 to perform preference judgments between the scores produced by \agentname{} and Single-round Directed Judging Pipeline following our predefined evaluation rubric. As illustrated in Figure~\ref{rubric}, for each generated image, we present human annotators with the original prompt, a set of key physical and logical principles that should be followed and the evaluation results from both \agentname{} and Directed Judging baseline. Annotators are then asked to choose which scoring system provides a more reasonable and accurate evaluation based on a detailed rubric that prioritizes the identification of core physical flaws and insightful reasoning. This pairwise preference task allows us to directly measure which evaluation method aligns better with expert-human judgment. Specifically, we ask gemini-2.5-pro to generate the set of key physical and logical principles behind the generated prompt.

\section{Evaluation Results of \benchname{}}
We presents more result showcases on \benchname{} via \agentname{} in Figure~\ref{examples}, including the internal output of World Knowledge Extractor(WKE), Hypothesis Formulator(HF), Visual Perceptor(VP) and Reasoning Judger(VR). In each part, we require it to return a dictionary in JSON format.

\onecolumn{}

\begin{algorithm*}[t]
\caption{Pseudo Code of \agentname{}}
\label{alg:evaluate-image}
\begin{algorithmic}[1]

\Require prompt $x$ (string), image $\mathcal{I}$, World Knowledge Extractor(WKE), Hypothesis Formulator(HF), Visual Perceptor(VP), Reasoning Judger(RJ)
\Ensure  PW-Score $S_{PW} \in [0,10]$

\Function{EvaluateImage}{$x$, $\mathcal{I}$}
    \State $\mathcal{E} \gets$ \Call{WKE}{$x$} 
        \Comment{$\mathcal{E}$: expectation\_list returned by World-Knowledge Extractor}
    \State $\mathcal{F} \gets [\,]$ 
        \Comment{$\mathcal{F}$: facts\_to\_check, an empty list of factual structures}

    \ForAll{$E_i \in \mathcal{E}$}
        \State $\mathcal{Q}_i \gets$ \Call{HF}{$E_i.\text{expectation}$} 
            \Comment{$\mathcal{Q}_i$: question\_list generated by Hypothesis Formulator}
        \State $\mathcal{A}_i \gets [\,]$ 
            \Comment{$\mathcal{A}_i$: reported\_answers initialized empty}

        \ForAll{$Q_{ij} \in \mathcal{Q}_i$}
            \State $A_{ij} \gets$ \Call{VP}{$\mathcal{I}$, $Q_{ij}$} 
                \Comment{$A_{ij}$: answer produced by Visual Perceptor}
            \State \Call{Append}{$\mathcal{A}_i$, $A_{ij}$}
        \EndFor

        \State $M_i \gets$ \Call{Merge}{$\mathcal{A}_i$} 
            \Comment{$M_i$: merged\_facts combining all reported answers}
        \State $f_i \gets \{E_i.\text{importance}, E_i.\text{expectation}, M_i, \mathcal{I}\}$ 
            \Comment{$f_i$: fact object representing a unit of evaluation}
        \State \Call{Append}{$\mathcal{F}$, $f_i$}
    \EndFor

    \State $\mathcal{S} \gets$ \Call{RJ}{$\mathcal{F}$} 
        \Comment{$\mathcal{S}$: score\_output returned by Referee Judge}
    \State $S_{1} \gets \mathcal{S}.\text{layer\_1\_adherence\_score}$
    \State $S_{2} \gets \mathcal{S}.\text{layer\_2\_realism\_score}$
    \State $S_{3} \gets \mathcal{S}.\text{layer\_3\_nuance\_score}$
    \State $S_{PW} \gets 0.25*S_{1} + 0.5*S_{2} + 0.25*S_{3}$
    \State \Return $S_{PW}$
\EndFunction

\end{algorithmic}
\end{algorithm*}

\end{document}